\documentclass[a4paper,pagenum,french]{rnti}
\usepackage{comment}
\usepackage{graphicx}
\usepackage[T1]{fontenc}
\usepackage[utf8]{inputenc}
\usepackage{amsmath, amssymb}
\usepackage{hyperref}
\usepackage{url}

\titrecourt{Extraction multi-étiquettes de relations en utilisant des couches de Transformer}

\nomcourt{ N. L. Le et G. Tagny Ngompe}

\titre{Extraction de relations multi-étiquettes en utilisant des modèles pré-entraînés et des couches de Transformer}

\auteur{Ngoc Luyen Le \affil{1}\affilsep\affil{2}\, Gildas Tagny Ngompé}

\affiliation{
    \affil{1}Gamaizer.ia, 93340 Le Raincy, France\\
    \affil{2}Université de technologie de Compiègne, CNRS, Heudiasyc \\(Heuristics and Diagnosis of Complex Systems), \\CS 60319 - 60203 Compiègne Cedex, France\\
 }
 
\resume{%
Nous présentons dans cet article le modèle \textbf{BTransformer18}, une architecture d'apprentissage profond conçue pour l'extraction de relations multi-étiquettes dans des textes en français. Notre approche combine les capacités de représentation contextuelle de modèles de langage pré-entraînés de la famille BERT, tels que BERT, RoBERTa, ainsi que leurs versions francophones CamemBERT et FlauBERT, avec la puissance des encodeurs Transformer pour capturer les dépendances à long terme entre les tokens. Les expérimentations menées sur le jeu de données du défi TextMine'25 montrent que notre modèle atteint des performances supérieures, en particulier en utilisant \textbf{CamemBERT-Large}, avec un score F1-macro de 0,654, surpassant les résultats obtenus avec \textbf{FlauBERT-Large}. Ces résultats démontrent l'efficacité de notre approche pour l'extraction automatique de relations complexes dans des rapports de renseignement. 
}
\summary{In this article, we present the BTransformer18 model, a deep learning architecture designed for multi-label relation extraction in French texts. Our approach combines the contextual representation capabilities of pre-trained language models from the BERT family—such as BERT, RoBERTa, and their French counterparts CamemBERT and FlauBERT—with the power of Transformer encoders to capture long-term dependencies between tokens. Experiments conducted on the dataset from the TextMine'25 challenge show that our model achieves superior performance, particularly when using CamemBERT-Large, with a macro F1 score of 0.654, surpassing the results obtained with FlauBERT-Large. These results demonstrate the effectiveness of our approach for the automatic extraction of complex relations in intelligence reports.
}

\begin{document}
\section{Introduction}
L'extraction de relations est une tâche fondamentale en traitement automatique du langage naturel (TALN), visant à identifier et classifier les relations sémantiques entre des entités nommées au sein d'un texte \citep{nasar2021named}. Dans le cas des rapports de renseignement, cette tâche revêt une importance particulière pour la structuration de l'information et la facilitation de l'analyse. Les approches traditionnelles reposent souvent sur des méthodes d'apprentissage supervisé nécessitant des ressources annotées considérables et peinent à capturer les relations complexes présentes dans les textes non structurés.

Le défi TextMine'25 \citep{defi-text-mine-2025} fournit un jeu de données précieux pour faire progresser la recherche dans ce domaine. Ce jeu de données est composé de 800 rapports de renseignement factices en français, annotés avec des mentions d'entités, leurs types, attributs et relations. Il pose un défi significatif en raison de la complexité des relations et de la nécessité de modèles capables de gérer la classification multi-étiquettes.

Avec l'émergence des modèles de langage pré-entraînés, tels que ceux de la famille BERT, des améliorations significatives ont été réalisées dans diverses tâches de TALN. BERT \citep{devlin2018bert} et RoBERTa \citep{liu2019roberta} ont établi de nouveaux standards pour la langue anglaise, tandis que leurs équivalents francophones, CamemBERT \citep{martin2019camembert} et FlauBERT \citep{le2019flaubert}, ont été développés pour traiter les spécificités de la langue française. Ces modèles, membres de la famille BERT, exploitent un pré-entraînement à grande échelle sur des corpus massifs pour produire des embeddings contextualisés riches.

Parallèlement, les encodeurs Transformer \citep{vaswani2017attention,wolf2020transformers} ont démontré une capacité exceptionnelle à modéliser les dépendances à long terme dans les séquences grâce aux mécanismes d'attention. Combiner les forces des modèles de langage pré-entraînés et des encodeurs Transformer offre une direction prometteuse pour améliorer les capacités d'extraction de relations.

Dans cet article, nous introduisons \textbf{BTransformer18}, un modèle qui intègre les modèles de langage pré-entraînés de la famille BERT avec des encodeurs Transformer pour l'extraction de relations multi-étiquettes dans des textes en français. Notre architecture comprend trois couches principales : \textbf{Embeddings Contextuels}, où nous obtenons les embeddings des tokens en utilisant des modèles comme CamemBERT et FlauBERT ; \textbf{Encodeurs Transformer}, qui capturent les dépendances complexes entre les tokens ; et \textbf{Agrégation et Classification}, où nous agrégeons les représentations et prédisons les relations.

Nous évaluons notre modèle sur le jeu de données du défi TextMine'25, en réalisant des expérimentations approfondies pour évaluer l'impact des différents modèles pré-entraînés sur la tâche. Nos résultats montrent que l'utilisation de \textbf{CamemBERT-Large} conduit à des gains de performance significatifs, atteignant un score F1-macro de 0,654, surpassant les modèles basés sur \textbf{FlauBERT-Large}.

Dans la section 2, nous détaillons notre approche proposée en décrivant l'architecture de \textbf{BTransformer18}. La section 3 présente le cadre expérimental, y compris le jeu de données, les étapes de prétraitement, les procédures d'entraînement, et des résultats de notre modèle. Enfin, dans la section 4, nous concluons et proposons des perspectives pour de futures recherches.

\section{Approche proposée}

\begin{figure}[h!]
    \centering
    \includegraphics[width=\linewidth]{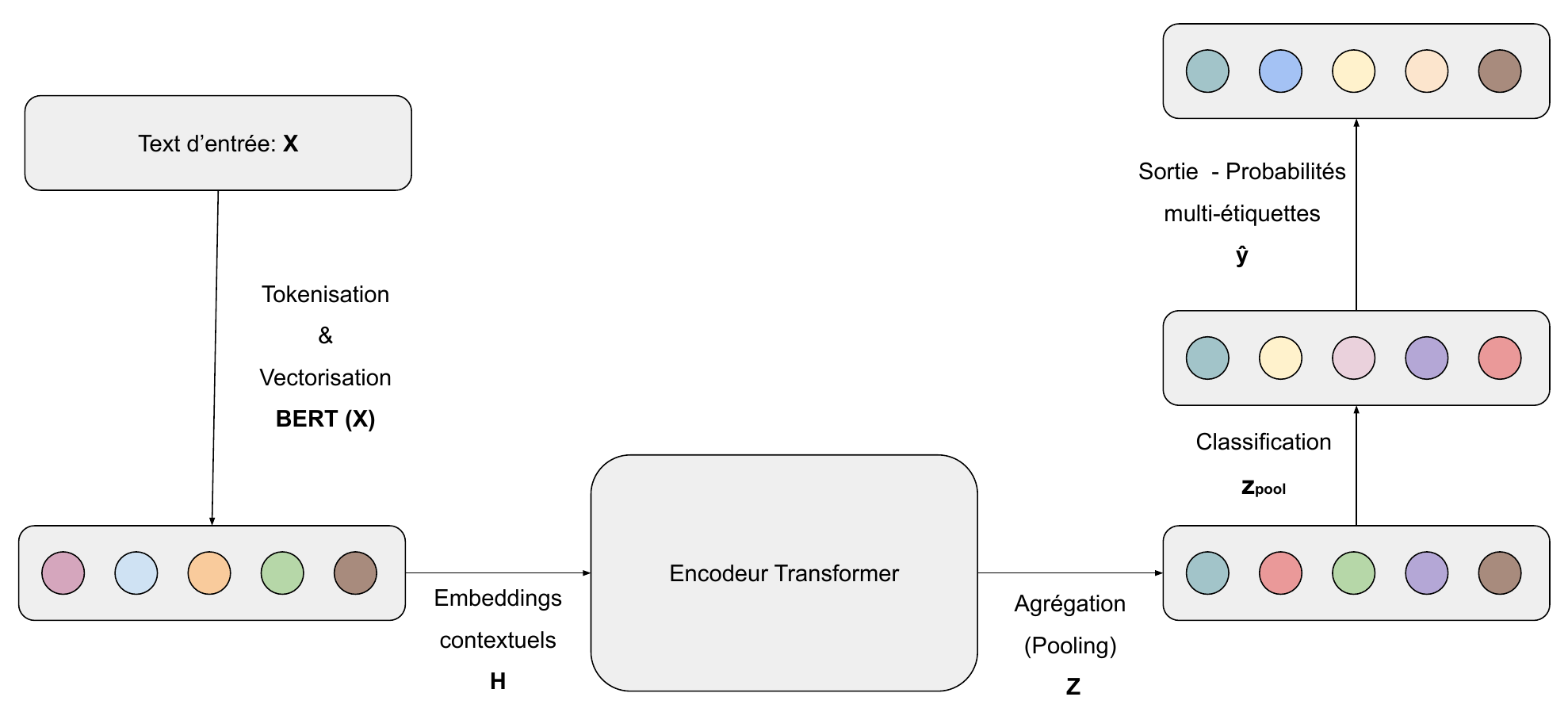}
    \caption{Représentation schématique de l'architecture \textbf{BTransformer18}. }
    \label{fig:architecture}
\end{figure}

Notre approche, illustrée dans la figure~\ref{fig:architecture}, s’appuie sur la logique de \textit{fine-tuning} d’un modèle de langage pré-entraîné, composée d’un \textit{body} (généraliste pour la langue ou le domaine) et d’une \textit{head} (spécifique à la tâche d’extraction de relations). Dans notre modèle, le \textit{body} est assuré par \textbf{CamemBERT-Large}, choisi pour sa spécialisation en français. La \textit{head} de classification inclut une couche cachée basée sur l’architecture Transformer \citep{vaswani2017attention}, permettant d’exploiter le mécanisme d’attention pour l’extraction de relations. La sortie est ensuite produite par une simple couche dense, qui assure la classification multi-label. 

\subsection{Embeddings contextuels}

Un des modèles de langage pré-entraînés de la famille BERT (par exemple, BERT, CamemBERT, FlauBERT, etc.) est employé afin d’obtenir les embeddings contextuels des tokens du texte d’entrée. Soit $\mathbf{X} = [x_1, x_2, \dots, x_T]$ la séquence de tokens, où $T$ est la longueur de la séquence. Ces  modèles de langage pré-entraînés génère pour chaque token $x_t$ un embedding contextuel $\mathbf{h}_t \in \mathbb{R}^d$, où $d$ est la dimension cachée du modèle, c’est-à-dire la taille du vecteur latent (par exemple, 768 pour la version BERT-base et 1024 pour la version BERT-Large).

\begin{equation}
\mathbf{H} = [\mathbf{h}_1, \mathbf{h}_2, \dots, \mathbf{h}_T] = \text{BERT}(\mathbf{X})
\end{equation}

Les embeddings contextuels $\mathbf{H}$ désignent les représentations vectorielles produites par un modèle de langage pré-entraîné  pour chaque \emph{token} du texte d’entrée. Contrairement aux \emph{word embeddings} classiques (tels que \texttt{word2vec} \citep{mikolov2013efficient} ou \texttt{GloVe} \citep{pennington2014glove}), qui associent à chaque mot un unique vecteur indépendant du contexte, les embeddings contextuels tiennent compte du voisinage lexical et syntaxique. Ainsi, le même mot peut avoir des représentations différentes selon son sens et sa position dans la phrase.

\subsection{Encodeurs transformer}

Les embeddings contextuels \(\mathbf{H}\) sont ensuite transmis à travers \(L\) couches d’encodeurs Transformer \citep{vaswani2017attention}, afin de capturer les dépendances à long terme entre les tokens. Cette architecture permet de modéliser efficacement la structure du texte et de faire émerger des représentations de plus en plus riches à mesure que les couches s’empilent. Chaque couche \(\ell\) transforme la représentation \(\mathbf{Z}^{(\ell-1)}\) en \(\mathbf{Z}^{(\ell)}\) comme suit :
\begin{equation}
\mathbf{Z}^{(\ell)}
= \text{TransformerEncoder}^{(\ell)} \Bigl(\mathbf{Z}^{(\ell-1)}\Bigr),
\quad \ell = 1, \dots, L,
\end{equation}
où \(\mathbf{Z}^{(0)} = \mathbf{H}\). Le fonctionnement de chaque couche peut être décomposé en quatre étapes principales : Attention multi-têtes, 1\textsuperscript{er}  Add \& Norm, Feed-forward positionnel et 2\textsuperscript{ème} Add \& Norm.

\paragraph{Attention multi-têtes:}
La première étape consiste à appliquer un mécanisme d’attention multi-têtes, noté \(\text{MultiHeadAttention}\), sur \(\mathbf{Z}^{(\ell-1)}\) :
\begin{equation}
\mathbf{A}^{(\ell)}
= \text{MultiHeadAttention}\Bigl(\mathbf{Z}^{(\ell-1)}\Bigr)
\end{equation}
Ce module scinde l’espace latent en plusieurs \emph{têtes} d’attention, permettant au modèle de se focaliser sur différents aspects du contexte. L’attention proprement dite est calculée comme suit :
\begin{equation}
\text{Attention}(\mathbf{Q}, \mathbf{K}, \mathbf{V})
= \text{softmax}\Bigl(\frac{\mathbf{Q}\mathbf{K}^\top}{\sqrt{d_k}}\Bigr)\mathbf{V},
\end{equation}
où \(\mathbf{Q}\), \(\mathbf{K}\) et \(\mathbf{V}\) sont respectivement les matrices des requêtes, des clés et des valeurs, et \(d_k\) la dimension des clés. Grâce à ce mécanisme, le modèle peut se focaliser sur diverses parties de la séquence, offrant ainsi une capture fine des dépendances sémantiques et syntaxiques.

\paragraph{1\textsuperscript{er} Add \& Norm:}
Une fois la matrice d’attention \(\mathbf{A}^{(\ell)}\) calculée, on l’additionne à l’entrée de la couche \(\mathbf{Z}^{(\ell-1)}\). Cette somme est ensuite normalisée :
\begin{equation}
\mathbf{U}^{(\ell)}
= \text{LayerNorm}\Bigl(\mathbf{Z}^{(\ell-1)} + \mathbf{A}^{(\ell)}\Bigr)
\end{equation}
L’addition permet de conserver les informations initiales (via le \emph{skip connection}), tandis que la normalisation de couche (\texttt{LayerNorm}) \citep{ba2016layer} stabilise l’apprentissage et facilite la convergence.

\paragraph{Feed-forward positionnel:}
Chaque position dans \(\mathbf{U}^{(\ell)}\) est ensuite traitée de manière indépendante par un réseau pleinement connecté, souvent appelé \emph{position-wise feed-forward network} (\(\text{FFN}\)). On calcule :
\begin{equation}
\mathbf{F}^{(\ell)} = \text{FFN}\Bigl(\mathbf{U}^{(\ell)}\Bigr)
\end{equation}
Ce réseau permet d’enrichir localement la représentation de chaque token, en transformant linéairement puis en appliquant une fonction d’activation non linéaire (par exemple, \(\text{ReLU}\)).

\paragraph{2\textsuperscript{ème} Add \& Norm:}
Enfin, on ajoute la sortie du \textit{feed-forward} \(\mathbf{F}^{(\ell)}\) à \(\mathbf{U}^{(\ell)}\), puis on normalise à nouveau :
\begin{equation}
\mathbf{Z}^{(\ell)}
= \text{LayerNorm}\Bigl(\mathbf{U}^{(\ell)} + \mathbf{F}^{(\ell)}\Bigr)
\end{equation}
À l’issue de cette étape, on obtient la sortie finale de la \(\ell\)-ième couche, qui sert d’entrée à la couche suivante. En répétant ces opérations \(L\) fois, l’encodeur Transformer parvient à agréger des informations à différentes échelles, améliorant ainsi sa capacité à modéliser les relations sémantiques et structurelles présentes dans la séquence d’entrée.

\subsection{Agrégation et classification}
La sortie du dernier encodeur Transformer est une séquence de représentations 
$
\mathbf{Z}^{(L)} 
= \bigl[ \mathbf{z}_1, \mathbf{z}_2, \dots, \mathbf{z}_T \bigr],$
où \(\mathbf{z}_t \in \mathbb{R}^d\) représente la représentation contextuelle du \(t\)-ième token. Pour réduire cette séquence à un unique vecteur reflétant l’information globale, nous effectuons une agrégation par moyenne~:
\begin{equation}
\mathbf{z}_{\text{pool}}
= \frac{1}{T} \sum_{t=1}^{T} \mathbf{z}_t
\end{equation}
Cette opération, appelée \emph{mean pooling}, permet de combiner les informations contenues par chaque token, tout en préservant la structure sémantique globale de la séquence. Elle est souvent moins sensible au bruit que d’autres méthodes d’agrégation (telles que l’utilisation d’un jeton spécial \texttt{[CLS]}) et tend à mieux lisser les variations locales.

\medskip

Une fois la représentation globale \(\mathbf{z}_{\text{pool}}\) obtenue, nous employons une couche de classification pour prédire les relations associées à la séquence. Plus précisément, nous appliquons une transformation linéaire, puis une fonction sigmoïde :
\begin{equation}
\hat{\mathbf{y}}
= \sigma\Bigl( \mathbf{W} \,\mathbf{z}_{\text{pool}} + \mathbf{b} \Bigr),
\end{equation}
où \(\mathbf{W} \in \mathbb{R}^{C \times d}\) et \(\mathbf{b} \in \mathbb{R}^C\). Ici, \(C\) désigne le nombre de classes de relations, et \(\sigma\) est la fonction sigmoïde appliquée à chaque composante. La sortie \(\hat{\mathbf{y}} \in [0,1]^C\) est alors interprétée comme un vecteur de probabilités indiquant la présence ou l’absence de chaque relation.

\section{Expérimentations et résultats}
Dans cette section, nous décrivons les expérimentations menées pour évaluer les performances de notre modèle \textbf{BTransformer18} sur la tâche d'extraction de relations multi-étiquettes. Nous présentons d'abord le jeu de données utilisé, puis les détails de l'implémentation et des paramètres d'entraînement. Enfin, nous discutons des résultats obtenus.
\subsection{Jeu de données et prétraitement des données}
Le jeu de données utilisé pour les expérimentations est constitué de 800 rapports de renseignement factices fournis dans le cadre du défi TextMine'25 \citep{defi-text-mine-2025}. En général, les données d'entrée se composent d'un texte, des mentions et des types des entités ainsi que des attributs associés.

Par rapport à la tokenisation : les textes sont tokenisés en utilisant le tokenizer associé au modèle de language pré-entraîné, garantissant une correspondance optimale avec les embeddings. Les entités annotées sont alignées avec les tokens pour former des paires d'entités potentielles $(e_i, e_j)$.

Sur la construction des Paires d'Entités: pour chaque document, nous générons toutes les paires possibles d'entités annotées, où $e_i$ et $e_j$ sont des entités du texte. Chaque paire est associée à un vecteur de labels multi-étiquettes $\mathbf{y}_{ij} \in \{0,1\}^C$, indiquant les relations existantes entre $e_i$ et $e_j$.

\subsection{Entraînement du modèle}
\paragraph{Fonction de perte :}
La Binary Cross-Entropy (BCE) est utilisée comme fonction de perte pour la classification multi-étiquettes. Cette fonction est définie par l'équation suivante :
\begin{equation}
\mathcal{L} = - \frac{1}{N} \sum_{n=1}^{N} \sum_{c=1}^{C} \left[ y_{nc} \log(\hat{y}_{nc}) + (1 - y_{nc}) \log(1 - \hat{y}_{nc}) \right],
\end{equation}
où \(N\) représente le nombre total de paires d'entités dans le lot d'entraînement, \(y_{nc} \in \{0, 1\}\) est l'étiquette réelle pour la classe \(c\), et \(\hat{y}_{nc} \in [0, 1]\) est la probabilité prédite par le modèle pour cette classe. 

\paragraph{Optimisation :}
L'optimisation du modèle est réalisée à l'aide de l'optimiseur \textbf{AdamW} \citep{loshchilov2017decoupled}, avec un taux d'apprentissage initial \(\alpha_0 = 2 \times 10^{-5}\). Un scheduler avec \textit{warm-up} est appliqué pour ajuster dynamiquement le taux d'apprentissage au cours de l'entraînement. La mise à jour du taux d'apprentissage est donnée par :
\begin{equation}
\alpha_t = \alpha_0 \times \min\left( \frac{t}{t_{\text{warmup}}}, 1 \right),
\end{equation}
où $t$ est le nombre d'itérations effectuées, et $t_{\text{warmup}}$ est le nombre d'itérations pendant la phase de warm-up.

\paragraph{Régularisation :}
Pour réduire le risque de surapprentissage, plusieurs techniques de régularisation sont intégrées dans l'architecture. Un \textbf{dropout} \citep{srivastava2014dropout} avec un taux de \(p = 0.1\) est appliqué dans les couches Transformer et la couche de classification. En parallèle, une régularisation \textbf{L2} (pondération de décroissance) est utilisée sur les poids du modèle pour limiter leur amplitude et améliorer la généralisation.
\begin{table}[h]
\centering
\label{tab:hyperparameters}
\begin{tabular}{p{7cm} p{5.5cm}}
\hline
\textbf{Hyperparamètre}                & \textbf{Valeur}                   \\ \hline
Modèle de base                & CamemBERT-Large                   \\ 
Taille des embeddings         & 1024                              \\ 
Nombre de couches Transformer & 2                                 \\ 
Nombre de têtes d'attention   & 8                                 \\ 
Taux de dropout               & 0.1                               \\ 
Longueur maximale de séquence & 150                               \\ 
Taille du batch (entraînement) & 16                                \\ 
Taille du batch (validation)  & 16                                \\ 
Taux d'apprentissage initial (\(\alpha_0\)) & \(2 \times 10^{-5}\)             \\ 
Optimiseur                    & AdamW                             \\ 
Scheduler                     & Linear avec warm-up               \\ 
Proportion du warm-up         & 10\%                              \\ 
Nombre maximal d'époques      & 50                                \\ 
Patience pour l'arrêt anticipé & 3                                 \\ 
Classes de classification     & 37 (multi-étiquettes)             \\ \hline
\end{tabular}
\caption{Hyperparamètres utilisés pour l'entraînement du modèle \textbf{BTransformer18}.}
\end{table}

\paragraph{Évaluation:}
Les performances du modèle sont mesurées à l'aide de métriques spécifiques à la classification multi-étiquettes. En particulier, la F1-mesure macro est employée pour évaluer la précision et le rappel moyens sur toutes les classes, pondérant chaque classe de manière égale \citep{defi-text-mine-2025}. 

\subsection{Résultats}


Les résultats obtenus par notre modèle sont présentés dans le tableau \ref{tab:results}, avec une comparaison entre deux modèles de langage pré-entraînés pour le français. Les expérimentations ont été réalisées en utilisant le modèle \textbf{BTransformer18}, en suivant les configurations et les hyperparamètres décrits dans la table \ref{tab:hyperparameters}.

\begin{table}[h]
    \centering
    \begin{tabular}{p{11cm} p{1.5cm}}
        \hline
        \textbf{Modèle} & \textbf{Score} \\ 
        \hline
        BTransformer18 (CamemBERT-Large) &  0,654 \\ 
        BTransformer18 (FlauBERT-Large) & 0,620 \\ 
        \hline
    \end{tabular}
    \caption{Résultats des modèles sur le jeu de données du défi TextMine'25.}
    \label{tab:results}
\end{table}
\begin{figure}[h!]
\begin{center}
 \includegraphics[width=13cm]{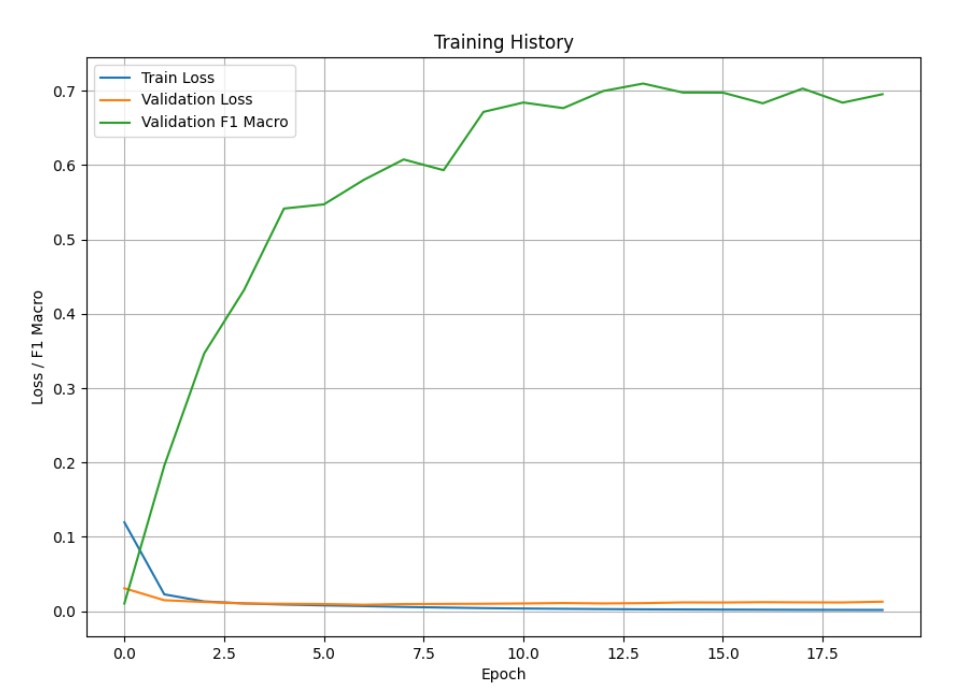}
 \caption{Évolution des pertes et de la F1-mesure macro de validation pour le modèle \textbf{BTransformer18} avec \textbf{CamemBERT-Large}.} \label{fig_exemple}
\end{center}
\end{figure}

Les résultats montrent que le modèle \textbf{BTransformer18} utilisant \textbf{CamemBERT-Large} obtient un score de \textbf{0,654}, tandis que celui utilisant \textbf{FlauBERT-Large} atteint un score de \textbf{0,620}, indiquant une amélioration de \textbf{3,4 points de pourcentage} grâce à l'intégration de \textbf{CamemBERT-Large}. Comme illustré dans la figure \ref{fig_exemple}, les courbes d'entraînement de \textbf{BTransformer18} avec \textbf{CamemBERT} montrent une convergence rapide des pertes d'entraînement et de validation au cours des premières époques, suivie d'une stabilisation. Simultanément, la F1-mesure macro de validation progresse rapidement avant d'atteindre un plateau, indiquant une amélioration significative des performances de classification multi-étiquettes. L'écart modéré entre les pertes d'entraînement et de validation met en évidence une bonne généralisation du modèle, sans signe de surapprentissage, même après plusieurs époques. Ces résultats soulignent l'efficacité de \textbf{BTransformer18} à exploiter les représentations contextuelles riches de \textbf{CamemBERT-Large}, permettant d'apprendre des relations complexes tout en maintenant des performances stables sur les données de validation.

Les résultats suggèrent que le choix du modèle de langage pré-entraîné a un impact significatif sur les performances de l'extraction de relations multi-étiquettes. L'utilisation de \textbf{CamemBERT-Large} semble offrir une meilleure représentation contextuelle pour les textes du défi, ce qui se traduit par une amélioration notable des performances.

Pour garantir la reproductibilité des résultats et permettre une exploration plus approfondie, le code source complet de l'implémentation est disponible publiquement sur le dépôt GitHub suivant : \url{https://github.com/lengocluyen/relation_extraction_textmine25}. 

\section{Conclusion et perspectives}
Dans cet article, nous avons présenté le modèle \textbf{BTransformer18}, une architecture combinant des modèles de langage pré-entraînés francophones, tels que \textbf{CamemBERT-Large} et \textbf{FlauBERT-Large}, avec des couches Transformer pour l'extraction de relations multi-étiquettes dans des rapports de renseignement. Les résultats expérimentaux ont démontré la supériorité de \textbf{CamemBERT-Large}, qui a obtenu un score F1 macro supérieur à celui de \textbf{FlauBERT-Large}. L'analyse des courbes d'entraînement a mis en évidence une convergence rapide et une bonne généralisation, montrant l'efficacité du modèle pour capturer des relations complexes tout en évitant le surapprentissage. En exploitant les avancées récentes en traitement du langage naturel, notre modèle démontre sa capacité à relever le défi de l'extraction automatique de relations complexes, tout en maintenant une classification précise grâce aux couches Transformer.

Bien que les résultats obtenus soient prometteurs, plusieurs axes d'amélioration peuvent être explorés dans de futurs travaux. Tout d'abord, l'enrichissement des données avec des corpus supplémentaires ou annotés dans d'autres domaines pourrait renforcer la robustesse du modèle et améliorer sa généralisation. Par ailleurs, l'incorporation de graphes de connaissances ou l'utilisation de modèles d'apprentissage par graphes pourrait améliorer la modélisation des relations complexes entre entités. De plus, l'intégration de grands modèles de langage (\textit{Large Language Models}, LLMs), comme GPT \citep{achiam2023gpt}, Mistral \citep{jiang2023mistral}, LLama \citep{touvron2023llama}, et des autres, pourrait offrir des représentations contextuelles encore plus riches et dynamiques, notamment pour des relations subtiles ou rares. Enfin, des techniques d'augmentation des données, combinées à des méthodes de régularisation avancées, ainsi qu'une optimisation des ressources computationnelles, permettraient de développer des modèles plus robustes et efficaces, adaptés à des applications en temps réel ou dans des environnements à ressources limitées. Ces pistes visent à élargir l'applicabilité du modèle tout en améliorant ses performances dans des tâches d'extraction de relations complexes.

\bibliographystyle{rnti}
\bibliography{references}

\end{document}